\documentclass{llncs}
\usepackage[utf8]{inputenc}
\usepackage{graphicx}
\usepackage{tabularx} 
\usepackage{multirow}
\usepackage{cite}
\usepackage{epstopdf}
\usepackage{longtable}
\usepackage{textcomp}
\usepackage[round]{natbib}

\begin{document}
\title{Supervised Term Weighting Metrics for Sentiment Analysis in Short Text}
\author{Hussam Hamdan*,**, Patrice Bellot*, Frederic Bechet**}
\institute{*Aix Marseille Université, CNRS, ENSAM, Universit{é} de Toulon, LSIS UMR 7296,13397, Marseille, France {hussam.hamdan,patrice.bellot}@lsis.org \\ *** Aix Marseille Université, CNRS, LIF UMR 7279, Marseille, France {frederic.bechet}@lif.univ-mrs.fr}

\maketitle
\begin{abstract}
Term weighting metrics assign weights to terms in order to discriminate the important terms from the less crucial ones. Due to this characteristic, these metrics have attracted growing attention in text classification and recently in sentiment analysis. Using the weights given by such metrics could lead to more accurate document representation which may improve the performance of the classification. While previous studies have focused on proposing or comparing different weighting metrics at two-classes document level sentiment analysis, this study propose to analyse the results given by each metric in order to find out the characteristics of good and bad weighting metrics. Therefore we present an empirical study of fifteen global supervised weighting metrics with four local weighting metrics adopted from information retrieval, we also give an analysis to understand the behavior of each metric by observing and analysing how each metric distributes the terms and deduce some characteristics which may distinguish the good and bad metrics. The evaluation has been done using Support Vector Machine on three different datasets: Twitter, restaurant and laptop reviews.
\end{abstract}

\section{Introduction}
Polarity classification is the basic task in sentiment analysis in which the polarity of a given text should be determined, i.e. whether the expressed opinion is positive, negative or neutral. This analysis can be done at different levels of granularity: Document Level, Sentence Level or Aspect Level.
Different machine learning approaches have been proposed for accomplishing this task: lexicon-based and supervised. The supervised methods have been widely used and have achieved good results since 2002.

Document representation is a critical component in sentiment analysis just like in information retrieval and text classification, the Vector Space Model (VSM) is one of the most popular models in which each document can be seen as a vector of independent features or terms, and each term assigned a weight according to a weighting schema. The basic weighting schema uses binary weights (w = 1 if the term is present in the document, and w = 0 if not). A better and much referenced weighting schema is the tf (Term Frequency) or the tf*idf (Inverted Document Frequency) schema. Many other schemas have been proposed aiming at making the text classifiers more accurate.

In sentiment analysis, the early work by \cite{pang_thumbs_2002} reported that binary weight schema outperforms the term frequency. More recent research has focused on more complex term weighting schemas which are usually called supervised weighting metrics as they exploit the categorical information. Some metrics have been adopted from information retrieval such as DeltaIDF \citep{martineau_delta_2009,paltoglou_study_2010}, later on several metrics have been proposed involving those adopted from information theory and widely used in text classification such as information gain and Mutual Information \citep{deng_study_2014}.
Recently, \cite{wu_reducing_2014} also tested several methods adopted from information retrieval and information theory, they also proposed a new metric called natural entropy (ne) inspired from information theory.

\noindent Just like Information Retrieval, term weight is depending on three factors:
\begin{enumerate}
 \item Local factor: which is a function of term frequency tf within the document.
 \item Global factor: which is a function of term frequency at the corpus level such as the document frequency.
 \item Normalization factor: which normalizes the weights in each document, the normalization can also be done for local and global factors. 
\end{enumerate}
This general definition of term weight is used in \citep{paltoglou_study_2010}. While \cite{wu_reducing_2014,deng_study_2014} considered that a supervised term weighting schema based on two basic factors: Importance of a Term in a Document (ITD) and Importance of a Term for expressing Sentiment (ITS), the ITD is exactly the local factor, ITS is the global factor in the general definition of term weighting.

We can also distinguish between the unsupervised weighting methods which only use the distribution of the term in the corpus for global weight without any category information just like in information retrieval, and supervised weighting methods which use the available category information for more efficient estimation of term importance. Thus, for each term in each category we get a score, the final score should be a function of these scores such as the maximum score, the sum or the weighted sum of the term scores over the categories.

This study presents an empirical study of four local weighting schema and fifteen supervised global weighting schema. Theses metrics are evaluated on three datasets provided in SemEval tasks: Sentiment Analysis in Twitter \citep{nakov_semeval-2013_2013} and Aspect-Based Sentiment Analysis \citep{pontiki_semeval-2015_2015}. 
In context of sentiment analysis, several studies have evaluated some schemas, but they all evaluated their schema using binary classification (if a given text is positive or negative) and at document level (movie reviews), they reported the results on different datasets but they did not explain the results or the behavior of each metric in order to understand why their proposed metrics improve the performance. In all these studies, Support Vector Machine (SVM) classifier has been used for evaluating the metrics.

The intuition beyond using these metrics is that a supervised weighting schema may give more realistic representation of a document which may improve the performance of a classifier.

The remaining of this study is organized as follows. Section 2 shows our research objectives. Section 3 outlines existing work concerning supervised weighting metrics in sentiment analysis. Section 4 describes the term weighting metrics. Datasets are presented in Section 5. Our experiments and analysis are discussed in section 6, and the conclusion and future work is presented in Section 7.

\section{Research Objectives}
Different term weighting metrics have been proposed and implemented for improving the performance in text classification. To our knowledge, all these studies have either compared some existing metrics or proposed a new one. But they have not analysed these metrics, they have not provide a mathematical analysis or a statistical analysis which may explain why a certain metric can be more useful than the others. In this study, we aim to understanding why some metrics give good results while others do not. The mathematical analysis cannot provide us with any reasonable explanation. It seems that all metrics depends on the same components but each metric constitues them with its own method. Therefore, we propose to analyse the metrics statistically, we study how each metric distributes the words in the corpus and try to deduce some characteristics of the metrics which give good and bad results. Thus, given a metric and a corpus we can estimate if it is good or bad according to how it distribute the corpus. 
For this purpose, we study the impact of term weighting on short text sentiment analysis (sentence or aspect level), and on three-classes classification problem (positive, negative, neutral). We formulate our study to address three questions:
\begin{enumerate}
 \item Are the global weighting metrics useful for sentiment analysis?
 
 \item If a global weighting metric is useful, are the local metrics useful?
 
 \item What does make a global metric useful and how can we interpret its performance?
\end{enumerate}
\section{Related Work}

Term weighting is the task of assigning scores to terms, these scores measure the importance of a term according to the objective task. A lot of term weighting methods have been proposed for Information Retrieval, all based on Salton's definition \citep{salton_term-weighting_1988} where term weight is function of three factors: term frequency, inverse document frequency, and normalization.

While the term weighting methods in Information Retrieval are unsupervised, many supervised methods have been proposed in Text Classification, they have proved their efficiency in many studies \citep{sebastiani_machine_2002,debole_supervised_2003,ren_class-indexing-based_2013, forman_extensive_2003, savoy_feature_2013}.

Supervised classification methods have been widely used for sentiment analysis, early work by \cite{pang_thumbs_2002} reported that SVM outperforms other classifiers and the binary term representation also outperforms term frequency. Thus, following research has used the binary representation. Recently, research has focused on more efficient term weighting methods to improve the performance of sentiment analysis, \cite{martineau_delta_2009} proposed Delta tf*idf in which the final term weight is the difference between tf*idf in positive class and negative class, their experiments have been done on two-class classification. Later on, \cite{paltoglou_study_2010} studied some variants of the classic tf*idf schema adapted to sentiment analysis which provides significant improvement in terms of accuracy. \cite{deng_study_2014} presented several supervised weighting metrics adopted from information theory and TC. \cite{wu_reducing_2014} reviewed several existing weighting metrics in SA, they found that existing methods suffer from over-weighting, thus they proposed two regularization techniques, singular term cutting and bias term in addition to a new weighting metric called natural entropy (ne) adopted from information theory. 
 
Many other researches in Sentiment Analysis and Text Classification have studied some weighting metrics for feature selection, their objective is to reduce the feature space size by selecting the most important features. Document Frequency Difference was proposed to automatically identify the words which are more useful for classifying sentiment \citep{farzindar_comparison_2010}. \cite{savoy_feature_2013} combined different weighting metrics to select the most important features. \cite{rehman_relative_2015} have proposed a new feature ranking metric termed as relative discrimination criterion (RDC), which takes document frequencies for each term count of a term into account while estimating the usefulness of a term.
\cite{haddoud_combining_2016} have studied 96 term-weighting metrics, and among them 80 metrics have not been used. A combination method with different metrics has been shown to get better performance. 

\section{Term Weighting Metrics}

In this section, we describe the weighting metrics for sentiment analysis. Firstly, Let us denote the set of documents by $D=\{d_{1},d_{2},..,d_{n}\}$, the set of classes $C=\{c_{1},c_{2},..,c_{m}\}$, $F = \{f_{1}, f_{2}, . . ., f_{r}\}$ is the vocabulary, which is set of terms in D. The document $d_{j}$ is represented by a bag-of-words vector: $d_{j} = (w_{1j}, w_{2j}, . . ., w_{lj})$, where wij stands for the weight of feature $f_{i}$ in document $d_{j}$. 
In our weighting methods, $w_{ij}$ is defined as function of three factors: local weight, global weight and normalization factor. The final weight will be the product of these three components:

$$wij = local_weight * global_weight * normalization$$

\subsection{Local Weight}
Local term weight is derived from the frequency of the term within the document. Table 1 shows four local weighting metrics which have been proposed in context of Information Retrieval.
\begin{table}[h!]
\begin{center}
\begin{tabularx} {\textwidth}{|l|c|X|}

   \hline
    Local weight & Notation & Formula\\
     \hline
  
      1 if tf$>$0; 0  otherwise & tp & term presence 1 for presence, 0 for absence\\
       \hline
      tf & tf &raw term frequency\\
       \hline
      $k+(1-k)\frac{tf}{max_{tf}}$ &atf &
Augmented term frequency, $max_{tf}$ is the maximum frequency of any term in the document,k is set to 0.5 for short documents.\\
 \hline
      $log(tf+1)$ &  logtf & Logarithm of raw term frequency\\

     \hline
  
\end{tabularx}

\caption{Local weighting metrics.}
\end{center}
\end{table}
\subsection{Global Weight}
Global term weighting metrics take into account the frequency of term in the whole collection, these metrics fall into two categories. The first one, known as unsupervised term weighting method, does not take category information into account. The second called supervised term weighting method which exploits the category information of training documents in the classification tasks. Before describing the fifteen supervised weighting metrics used in this study, we present the notation used to define these metrics, we use the words \textit{term} and \textit{feature} interchangeably:

$N$ : Number of documents in the corpus.

$N_{c}$ : Number of documents in the class \textit{c}.

$N_{\bar c}$ : Number of documents out of the class \textit{c}.

$df$ : Number of documents containing the feature \textit{f} in whole corpus.

$df_{c}$ : Number of documents in class c containing the feature \textit{f}.

$df_{\bar c}$: Number of documents out of class \textit{c} containing the feature \textit{f}.

$p(c)$ : The probability of the class \textit{c}.

$p(c|f)$ : The probability of class \textit{c} given the feature \textit{f}.

$p(f)$ : The probability of the feature \textit{f} in the corpus.

$d\bar f$ : The number of documents which do not contain the feature \textit{f} in the corpus.

$p(\bar c|f)$: The complimentary probability of class \textit{c} given the feature \textit{f}.

$p(c,f)$ : The joint probability of the class \textit{c} and the feature \textit{f}.

$p(c,\bar f)$: The probability of class \textit{c} in the documents which do not contain the feature \textit{f}.

$d\bar f \bar c$ : The number of documents out of \textit{c} that do not contain the feature \textit{f}.

$tf_{c}$ : The frequency of feature \textit{f} in the class \textit{c}.

$d \bar f c$ : The number of document do not contain the feature \textit{f} in the class \textit{c}.

\begin{table}[h!]
\begin{center}

\begin{tabularx} {\textwidth}{|l|c|X|}

    \hline
    Global Weighting Metric & Notation & Formula\\
     \hline
  
 Delta Smoothed IDF &  dsidf &\multirow{2}{*}{$dsidf(f,c)=log(\frac{N_{c} . df_{\bar c} +0.5} {N_{\bar c}. df_{c}+0.5} )$}\\
 &&\\
       \hline
      Delta BM25 IDF&dbidf&$dbidf(f,c)=log(\frac{(N_{c}-df_{c}+0.5).df_{\bar c}+0.5}  {(N_{\bar c} -df_{\bar c}+0.5).df_{c}+0.5} )$\\
       &&\\
       \hline
    Relevance Frequency & rf & $rf(f,c)=log( 2+\frac {dfc} {max(1,df_{\bar c})} )$\\
     &&\\
 \hline
    Information Gain & ig&$ ig(f)=-\sum_{c \in C}p(c).log(p(c))+
(\frac {df}  {N} .\sum_{c \in C}p(c | f) log(p(c | f))) +
(\frac {d \bar f}  {N}.sum_{c \in C}p(c | \bar f).log(p(c | \bar f)))$ \\
 &&\\

     \hline
 Pairwise Mutual Information & pmi&$pmi(f,c)=log(\frac{p(c,f)}  {p(c).p(f)})$\\
  &&\\
  \hline
 Natural Entropy & ne &$ne(f,c)=1+(p(c | f). log(p(c | f))+p(\bar c | f). log(p(\bar c | f)))$\\
  &&\\
    \hline
 Chi Square $x^{2}$  & chi & $x^{2}(f,c)= \frac{N.((df_{c} . d \bar f \bar c)-(df \bar c . d \bar f c))^{2}}  {df . d\bar f . N_{c} . N_{\bar c}} $\\
  &&\\
    \hline
 NGL Coefficient & ngl & $ngl(f,c)=\frac {\sqrt{N} (df_{c} . d \bar f_{\bar c} - df_{\bar c} . d \bar f_{c})} {\sqrt{df . d \bar f . N c . N \bar c} } $\\
  &&\\
    \hline
  Class Discrimination Measure & cdm & $ cdm(f,c)=|log \frac {p(f | c)}  {p(f | \bar c)} |$\\
   &&\\
    \hline
 Class Discrimination Measure  & cdm &  $cpd(f,c)=\frac {df_{c} -df_{ \bar c} } {df} $\\
  &&\\
    \hline
  Multinomial Z Score & zd & $zd(f,c)= \frac {tf_{c} - p(f).N_{c}}  {\sqrt{N_{c} . p(f) . (1-p(f))} }$\\
   &&\\
    \hline
 Kullback-Leibler Divergence & kl & $ kl(f,c)=p(c | f). log(\frac {p(c | f)}  { p(c)})$\\
  &&\\
    \hline
  Weighted Log Likelihood Ratio & wllr &  $wllr(f,c)=p(f | c).log(\frac {p(f | c)}  {p( f | \bar c)} )$\\
   &&\\
   \hline
  Odds Ratio & orr &$orr(f,c)=log(\frac {p(f | c). (1- p(f | \bar c))}  {p(f | \bar c).(1-p(f | c))} )$ \\
   &&\\
    \hline
   
\end{tabularx}
\caption{Global weighting metrics.}
\end{center}
\end{table}

\noindent The formulas of the fifteen weighting metrics are illustrated in Table 2:
\begin{enumerate}
  \item \textbf{BaseLine (bl)}\\
\textit{bl} assigns 1 to the feature existing in the document and 0 to absent features, this  binary representation is the baseline which we use to compare the efficiency of the other metrics.

\item \textbf{Delta Smoothed IDF (dsidf)}\\
\textit{dsidf} boosts the importance of terms that are unevenly distributed between one category and other categories and discounts evenly distributed words, the original version Delta TF-IDF is presented in \citep{martineau_delta_2009}, the smoothed version seems to be more efficient \citep{paltoglou_study_2010}.

\item \textbf{Delta BM25 IDF (dbidf)}\\
\textit{dbidf} is a variant of the \textit{dsidf} metric, BM25-IDF variant is used instead of classical IDF \citep{paltoglou_study_2010}.

\item \textbf{Relevance Frequency (rf)}\\
\textit{rf} boosts the terms which have high frequency in the positive category, that helps in selecting the positive samples from the negative ones \cite{lan_supervised_2009}.

\item \textbf{Information Gain (ig)}\\
\textit{ig} \cite{yang_comparative_1997} tries to find out how well each single feature separates the given dataset. Information entropy is used to measure the uncertainty of the feature and the dataset. \textit{ig} is the overall entropy of the training set minus the entropy of the feature. Thus, it measures how much the feature reduces the dataset uncertainty when it is observed.

\item \textbf{Pairwise Mutual Information (pmi)}\\
\textit{pmi} is a measure of association used in information theory and statistics, it measures how much the feature associates with the class \citep{church_word_1990}.

\item \textbf{Natural Entropy (ne)}\\
The basics beyond \textit{ne} \citep{wu_reducing_2014} is the more uneven the distribution of documents where a feature occurs, the larger the weight of this feature is. Thus, the entropy of the feature can express the uncertainty of the class given the feature. One minus this degree of uncertainty boosts the features those are unevenly distributed between the category and other categories.
 
\item \textbf{Chi Square $\chi ^{2}$ (chi)}\\
\textit{chi} \citep{yang_comparative_1997} measures the lack of independence between the feature and the category, the higher value of the $\chi ^{2}$, the closer relationship the feature and the class have.

\item \textbf{NGL Coefficient (ngl)} \\
\textit{ngl} is a variant of the Chi square metric \citep{ng_feature_1997}. I takes the squared root of Chi after replacing $d \bar f \bar c$ with $d \bar f_{\bar c}$ and $d \bar f c$ with $d \bar f_{c}$.

\item \textbf{Class Discrimination Measure (cdm)}\\
\textit{cdm} measures the difference between the distribution of the feature in one class and other classes \citep{chen_feature_2009}.

\item \textbf{Categorical Proportional Difference (cpd)}\\
\textit{cpd} is a ratio that considers the number of documents of a category in which the feature occurs and the number of documents from other categories in which the feature also occurs \citep{simeon_categorical_2008}.

\item \textbf{Multinomial Z Score (zd)}\\
\textit{zd} supposes that a feature follows binomial distribution, calculates Z transformation for a feature in each class, \textit{zd} boosts the highly unevenly distributed features among the classes, it gives high positive score for a feature in the class where it is highly frequent and negative score in the class where it rarely appears \citep{savoy_authorship_2012, hamdan_impact_2014}. 
 
\item \textbf{Kullback-Leibler Divergence (kl)}\\
\textit{kl} is a non-symmetric measure of the difference between the distribution of the category and the distribution of the category given the feature. A measure of how dissimilar the two distributions are. Useful feature value implies a high degree of dissimilarity.

\item \textbf{Weighted Log Likelihood Ratio(wllr)}\\
\textit{wllr} is a measure of how dissimilar are the distribution of the feature given the category and the distribution of the feature given the other categories \citep{nigam_text_2000}.

\item \textbf{Odds Ratio (orr)}\\
\textit{orr} gives a positive score to features that occur more often in one category than in the other, and a negative score if it occurs more in the other. A score of zero means the odds for a feature to occur in one category is exactly the same as the odds for it to occur in the other \citep{shaw_term-relevance_1995}. 
\end{enumerate}
\subsection{Normalization}
Document length normalization adjusts the term score in order to normalize the effect of document length on the document classification. The most known normalization factor in Information Retrieval is the cosine normalization in which each term score is divided by the square root of the sum of all squared term scores within the document.
$$cosine=\sqrt{w_{f1}^{2}+w_{f2}^{2}+...+w_fm^{2}} $$

\subsection{Score Aggregation}
For each term in the corpus, we need one score as a weight for this term. The majority of the previous global metrics produce one score in each class for each term. Therefore, we have to apply an aggregation function to get only one score. We can apply different aggregation functions like \textit{max}, \textit{min}, \textit{sum} or \textit{weighted sum}.

In this study, we choose the \textit{max} function which takes the maximum value of the term scores over classes, this function boosts the important terms in any class, if a term is important in positive class or negative or neutral it is more significant to give it a high score. One can think about other functions like the score of term in negative class, positive or neutral, but we think the \textit{max} is the best one as it has been widely applied. And the \textit{sum} function may be not efficient when the metrics produce negative and positive values, also the \textit{weighted sum}, which takes into account the probability of class, but this probability has already been considered within the majority of the metrics, and therefore that may result in undesirable redundancy. Thus, the final term score can be written as following:

$$score(f)=Max_{c \in C }(gmetric(f,c))$$

Where \textit{gmetric} is a global metrics which computes the correlation between the term f and the class c, it may be any one of the fifteen global metrics.

\section{Datasets}
In this section, we describe the three datasets which we used for evaluating the term weighting metrics. The first one is extracted from Twitter, the second and the third are composed of sentences extracted from restaurant and laptop reviews, respectively.
\subsection{Twitter Dataset}
This dataset consists of Twitter messages compiled for SemEval-2013 task 2 and SemEval-2014 task 9, the participants were provided with a script to download 10882 tweets annotated by their polarities (positive, negative, neutral), test set is provided for the SemEval workshop 2014. The task aims to determine whether the tweet of positive, negative, or neutral sentiment. The statistics about this dataset is shown in Table 3 \#pos, \#neg, \#neut, \#total, lAvg refer to the number positive documents, the number of negative documents, the number of neutral documents, the total number of documents, the average length of document (the average number of tokens in the document), respectively.
\subsection{Restaurant and Laptop Reviews Datasets}
The second dataset is extracted from restaurant and laptop reviews, provided by SemEval 2015 ABSA organizers \citep{pontiki_semeval-2015_2015} where each review is composed of several sentences and each sentence may contain several Opinion Target Expressions. The statistics about these datasets is shown in Table 3.

\begin{table}[h!]
\begin{center}
\begin{tabular}{|c||c|c|c|c|c|}

  \hline
   Dataset & Pos.& Neg. &neut.&total &lAvg \\
     \hline
  
       Twitter Train & 3640 &  1458 & 4586 & 9684 &  22 \\
       Twitter Test (2014) & 1015 & 242 & 682 & 1939 & 22\\

       Restaurant train & 1655 & 1198 & 403 & 53 & 16\\
       Restaurant test & 845 & 454 & 346 & 45 &16\\
       
       Lap train  &277 &1739 & 1973& 3989&17\\
      Lap test  & 173 & 761 & 949& 1883&17\\
     \hline
  
\end{tabular}
\end{center}
\caption{Statistics for Twitter, Restaurant and Laptop datasets.}
\end{table}

\section{Experiments}
\subsection{Experiment Setup}
We only use uni-grams as features without neither word stemming nor stop word removal for the three datasets, all terms are used whatever their occurrences in the corpus.  Support Vector Machine (SVM) is used as classifier, SVM has been widely used in sentiment classification because its performance exceeds other machine learners \citep{pang_thumbs_2002}.

We trained a L2-regularized L2-loss linear SVM using the implementation of LIBLINEAR \citep{fan_liblinear2008} where all parameters are set to their default values. L2-regularized and L2-loss give a higher performance than other regularization techniques and have been used in most previous studies. For each dataset, we first tokenized the text for obtaining the terms, then we assign one score for each term with each metric, four classifiers have been trained for each dataset with each global metric scores, i.e. one classifier for each local weighting schema with each global one.

We normalized all scores produced from each global metric to make their values in the interval [0,1], therefore we can compare them disregarding the different intervals and the effect of negative values given by some metrics.
For normalization, we applied the following formula:
$$score(t)=\frac {s-min} {max-min}$$
where \textit{s} is the original score produced by a global metric, \textit{min}, \textit{max} are the minimum, maximum value given by the metric, respectively.

\subsection{Experiment Evaluations}

\begin{table}[h!]
\begin{center}
\begin{tabular}{|c|c|c|c|c||c|c|c|}
  \hline
     & tp & tf & atf & logtf & sumstd & stdy& meany \\
    \hline
       bl&	60.08&	59.92&	60.67&	59.80 & 	 0.000 & 	 1.000 & 	 0.000 \\
       zd&	61.35&	61.99&	62.29&	61.80& 	 5.490 & 	 0.515 & 	 0.030 \\
       ig&	59.55&	59.30&	58.81&	59.73& 	 2.650 & 	 0.004 & 	 0.016 \\
       pmi&	60.28&	60.20&	60.90&	59.93 & 	 4.507 & 	 0.916 & 	 0.126 \\
       ne &	60.57&	61.07&	61.26&	60.85& 	 15.689 & 	 0.511 & 	 0.243 \\
       chi&	59.49&	58.92&	58.71&	59.29& 	 2.316 & 	 0.004 & 	 0.015 \\
       kl&	62.12&	61.94&	62.54&	62.13& 	 15.769 & 	 0.616 & 	 0.327 \\
       wllr&	62.29&	62.59&	\textbf{63.33}&	62.41& 	 1.085 & 	 0.369 & 	 0.009 \\
       orr&	61.07&	61.23&	62.07&	60.97& 	 6.168 & 	 0.645 & 	 0.149 \\
       dsidf&	62.06&	62.31&	62.41&	62.07 & 	 8.879 & 	 0.408 & 	 0.135 \\
       dbidf&	61.48&	61.36&	61.79&	61.34& 	 6.873 & 	 0.658 & 	 0.102 \\
       rf&	62.42&	62.61&	\textbf{63.00}&	62.89& 	 10.903 & 	 0.203 & 	 0.110 \\
       cdm&	62.44&	62.26&	\textbf{63.01}&	62.70 & 	 10.306 & 	 0.375 & 	 0.159 \\
       ngl&	61.35&	61.99&	62.29&	61.80& 	 5.490 & 	 0.515 & 	 0.030 \\
       cpd&	61.40&	61.07&	62.24&	61.02& 	 12.781 & 	 0.728 & 	 0.282 \\
    \hline 
\end{tabular}

\caption{F1-score for each combination of global and local metrics on the Twitter dataset, the first four columns contain the F1-score of each combination but the last three ones contain the statistics of each global metric distribution.}
\end{center}
\end{table}
\begin{enumerate}
  \item \textbf{Twitter Dataset}

Table 4 reports the F1-scores of all combinations of global and local weighting metrics in Twitter dataset. The baseline \textit{bl*tp}, which refers to the binary representation of tweets has obtained 58.64\%. We will firstly discuss if the global weighting metrics can improve the baseline, then for those global ones which seem to give good results, we discuss if the local metrics work well. 

The majority of global metrics improve the performance, \textit{cdm, rf, wllr, kl, dsidf} give more than 62\% while \textit{zd, orr, ngl, dbidf} give more than 61\%, \textit{pmi} and \textit{ne} are close to the baseline but a little better, the remaining metrics do not improve the results but also they are not too far from the baseline such as \textit{ig}, \textit{chi}.

Regarding the local metrics, for the baseline run, \textit{bl*atf} improves the results. For other global metrics which their \textit{tp} weighting improves the baseline, \textit{tf} often improves the \textit{tp} except \textit{pmi, dbidf, cdm}. \textit{logtf} also improves the \textit{tp} except with \textit{cpd, dbidf, pmi}, but the \textit{atf} weighting schema always gives the best results over all good metrics which improve the baseline.

Thus, we can conclude that  most metrics improve the performance with \textit{tp} local weighting, and those who improve the \textit{bl*tp} produce the best results when using the \textit{atf} local weighting. \textit{wllr*atf, cdm*atf, rf*atf} produce the best F1-score of 63.33\%, 63.01\%, 63\%, respectively.

The idea behind all global metrics is to discriminate the terms which are unevenly distributed over the classes, the \textit{max} aggregation function will take the max value produced by each metric for each term, and therefore the final score for each term is expected to be high if the term is important in one class else if it is not important in all classes its value will be small, and the terms which occur similarly in the corpus will have different values in function of a global metric because they have different distributions over the classes in the corpus.

For understanding the behavior of each global weighting metric and being able to interpret its performance in Twitter dataset, we analyzed the distribution of term scores in function of their frequencies in whole corpus. Thus, we illustrate in Figure 1 how each global metric distributes the terms, fifteen diagram are drawn, one for each global metric with the \textit{max} aggregation function. X-axis represents the frequencies in the corpus, while Y-axis represents the final score of each term. For example, the point (1, 0.5) refers to a term \textit{w} which is repeated one time in the corpus and has 0.5 in function of a global metric. 
\begin{figure}[h!]
  \centering
   \includegraphics[width=\linewidth, height=12cm]{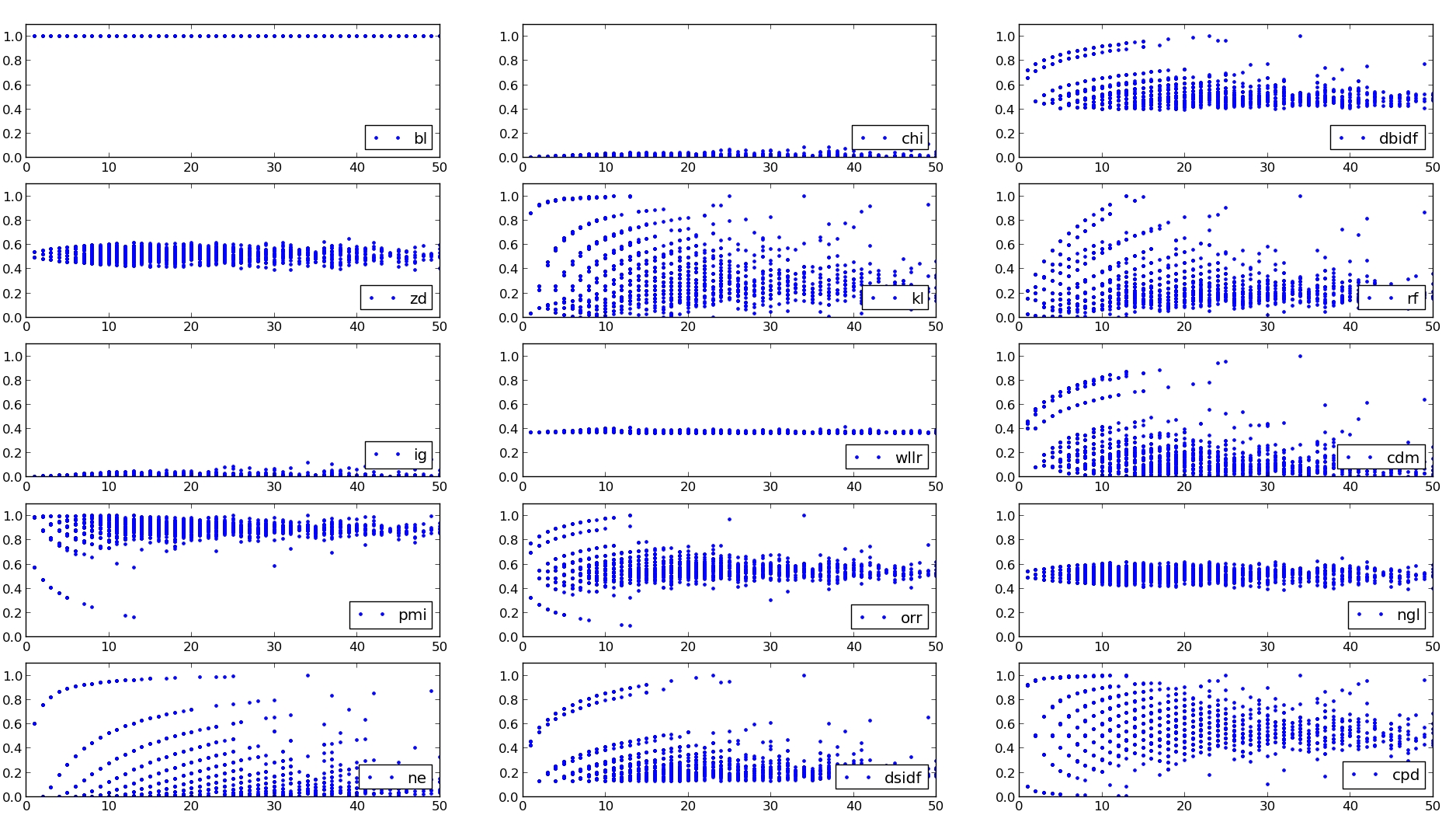}
  \caption{The distribution of each global metric in Twitter dataset.}
\end{figure}

As \textit{ig} and \textit{chi} have produced the lowest results, we observe their distributions firstly. It seems that they have a narrow distribution and the mean value for y-axis is close to Zero. Therefore, it is more probable that the narrow distribution is an indicator of an inefficient metric because it badly discriminates the terms and has several outliers. \textit{wllr} has also narrow distribution in spite of its good performance, but its mean value is far from zero. Thus, a metric with a narrow distribution and a mean value close to zero seems to be a bad one, therefore we can give an explanation of why the bias term has been efficient in \citep{wu_reducing_2014} where the authors found that adding a bias term to the score can improve the results and their results show that \textit{ne} without this regularization produces bad results in some datasets.

To provide some statistical evidences to our observations, we compute three measurements for each metric:
\begin{itemize}
 \item \textit{meany}: the mean value of the y-values produced by the metric. 
   \item \textit{sumstd}: we first compute the standard deviation for each frequency at X-axis in Figure 3.1. For example, the standard deviation of the terms which have the frequency of 8 in the corpus. Then, we sum the standard deviation of all frequencies. This \textit{sumstd} score designates how each global metric distributes the terms which have the same occurrence. If the metric produces scores very close and centered around the mean value, the \textit{sumstd} will be small.
    \item \textit{stdy}: the standard deviation of y-values produced by the metric which measures if the distribution is narrow or wide. If the metric produces scores very close and centered around the mean value, the \textit{sumstd} will be small. If the two measurements \textit{sumstd} and \textit{stdy} give small values, the distribution tends to be centered around the \textit{meany} value.
    
 \end{itemize} 

Table 4 shows for each metric the \textit{meany}, \textit{sumstd} and \textit{stdy}. We remark that \textit{wllr}, \textit{ig} and \textit{chi} have small values for textit{sumstd} and \textit{stdy} and therefore a distribution centered around their \textit{meany} .

\begin{table}[h!]
\begin{center}
\begin{tabular}{|c|c|c|c|c||c|c|c|}
    \hline
     & tp & tf & atf & logtf & sumstd & stdy& meany\\     
    \hline
     bl&	63.42&	65.37&	64.11&	64.48& 	 0.000 & 	 1.000 & 	 0.000 \\
     zd&	64.14&	66.27&	64.88&	65.41& 	 4.874 & 	 0.609 & 	 0.092 \\
     ig&	60.40&	61.17&	60.58&	61.02& 	 2.565 & 	 0.034 & 	 0.059 \\
     pmi&	63.12&	65.19&	64.49&	64.36& 	 3.344 & 	 0.875 & 	 0.155 \\
     ne &	63.20&	63.22&	62.96&	63.08& 	 6.016 & 	 0.729 & 	 0.160 \\
     chi&	61.55&	62.22&	61.21&	62.10& 	 2.180 & 	 0.030 & 	 0.058 \\
     kl&	65.06&	66.28&	65.13&	66.03& 	 11.146 & 	 0.561 & 	 0.339 \\
     wllr&	64.05&	65.24&	65.31&	64.33 & 	 3.226 & 	 0.330 & 	 0.046 \\
     orr&	63.87&	64.85&	64.10&	64.25& 	 5.139 & 	 0.601 & 	 0.156 \\
     dsidf&	62.42&	63.31&	62.78&	62.92& 	 6.733 & 	 0.509 & 	 0.139 \\
     dbidf&	63.84&	65.29&	63.33&	63.85& 	 5.585 & 	 0.636 & 	 0.129 \\
     rf&	64.69&	66.04&	65.62&	66.21& 	 7.223 & 	 0.208 & 	 0.153 \\
     cdm&	64.46&	64.70&	64.48&	64.50& 	 6.816 & 	 0.384 & 	 0.209 \\
     ngl&	64.14&	66.27&	64.88&	65.41& 	 4.874 & 	 0.609 & 	 0.092 \\
     cpd&	64.55&	66.80&	64.83&	65.31 & 	 8.300 & 	 0.656 & 	 0.346 \\
    \hline
\end{tabular}

\caption{F1-score for each combination of global and local metrics on the Restaurant dataset, the first four columns contain the F score of each combination but the last three ones contain the statistics of each global distribution.}
\end{center}
\end{table}

\item \textbf{Restaurant Dataset}

Table 5 shows the F1-scores of all combinations of local and global weighting metrics in Restaurant dataset. The baseline \textit{bl*tp}, which refers to the binary representation has obtained 63.42\%. Like with Twitter set, we will firstly discuss if the global weighting metrics can improve the baseline, then for those global ones which seem to give good results, we discuss if the local metrics work well. 

The majority of metrics improve the performance, \textit{kl} gives F1-score more than 65\%,\textit{ wllr, rf, cdm, ngl, cpd, zd} give more than 64\% while  \textit{orr, dbidf} give more than 63\% but they still higher than \textit{bl}, the other metrics decrease the performance however we should note that \textit{pmi} and \textit{ne} are still close to the baseline while \textit{ig, chi, dsidf} are lower by 2\% or 3\%.

Regarding local metrics, for the baseline, \textit{tf} produces the best result. For other metrics which improve the baseline, \textit{atf} often improves the \textit{tp} except with \textit{bl} and \textit{dbidf}. \textit{logtf} always improves but the \textit{tf} weighting schema always gives the best results over all metrics except rf.

Thus, the most metrics improve the performance with \textit{tp} local weighting, and those who improve the \textit{bl*tp} produce the best results when using the \textit{tf} weighting. \textit{cpd*tf, zd*tf, ngl*tf, rf*tf, rf*logtf} produce the best F1-score of 66.80\%, 66.27\%, 66.27\%, 66.04\%, 66.21\%, respectively.

Like for Twitter, Figure 3.2 shows the distribution of each metric over the dataset.
\begin{figure}[h!]
  \centering
   \includegraphics[width=\linewidth, height=12cm]{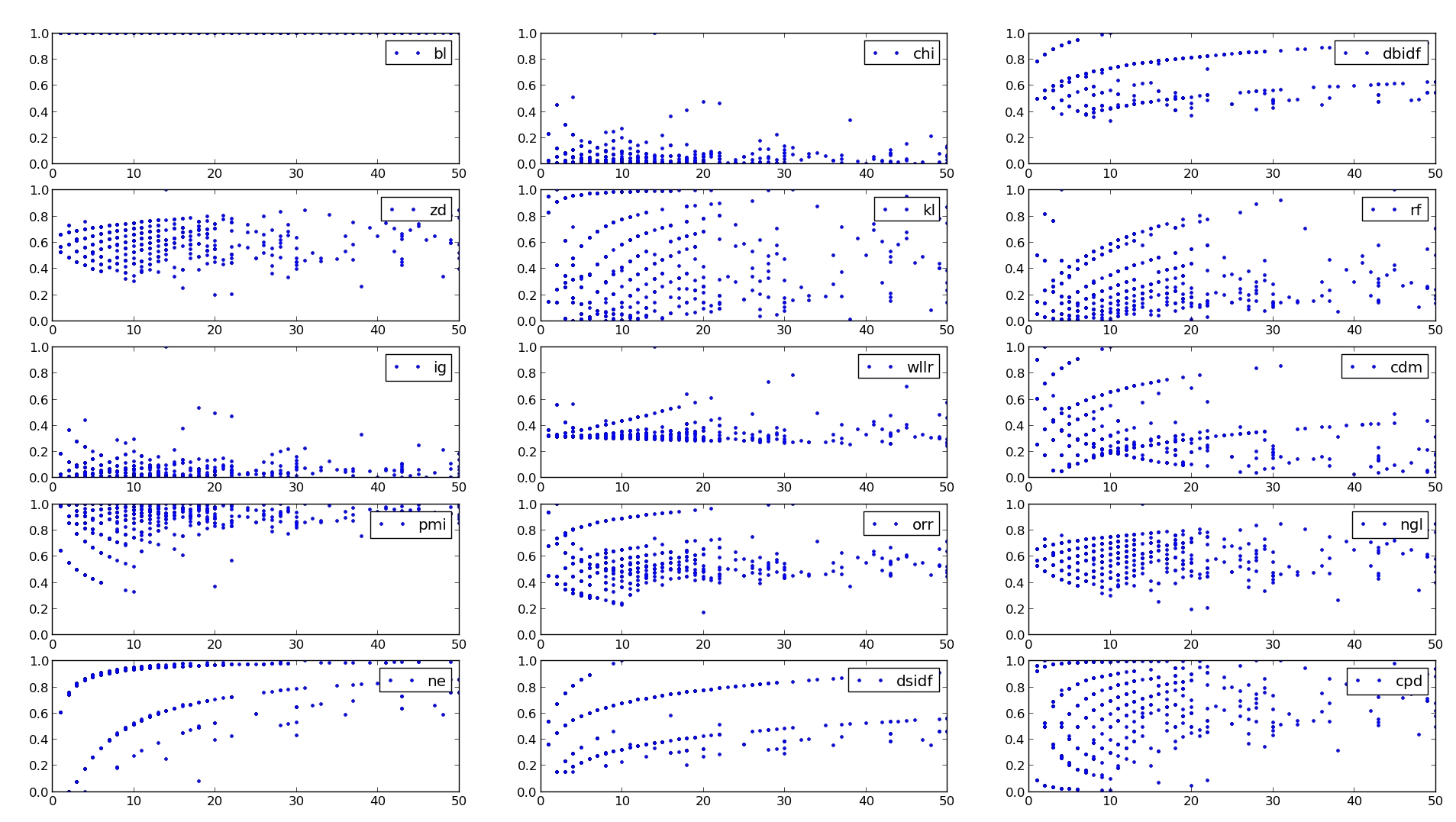}  
   \caption{The distribution of each global metric in Restaurant dataset.}
\end{figure}

\item \textbf{Laptop Dataset}

Table 6 displays the F1-scores of all combinations of local and global weighting metrics in laptop dataset. The baseline \textit{bl*tp}, which refers to the binary representation has obtained 69.33\%. Like with Twitter and Restaurant sets, we will firstly discuss if the global weighting metrics can improve the baseline, then for those global ones which seem to give good results, we discuss if the local metrics work well. 

The majority of metrics improve the performance, \textit{dsidf} and \textit{ne} give F1-score more than 71\%, \textit{dbidf} and \textit{rf} give more than 70\% while \textit{pmi, kl, wllr, orr, ngl and zd} give more than 69\% but they are still higher than \textit{bl}, the other metrics decrease the performance however we should note that \textit{ig, chi, cpd} are still close to the baseline.
\begin{table}[h!]
\begin{center}
\begin{tabular}{|c|c|c|c|c||c|c|c|}
    \hline
     & tp & tf & atf & logtf & sumstd & stdy& meany\\     
    \hline
          bl&	69.33&	69.72&	69.27&	69.46&           0.000 & 	 1.000 & 	 0.000 \\
          zd&	69.50&	69.70&	69.53&	69.69& 	 4.821 & 	 0.519 & 	 0.080 \\
          ig&	68.24&	68.23&	68.73&	68.18& 	 2.508 & 	 0.035 & 	 0.061 \\
          pmi&	69.42&	69.21&	69.07&	69.33& 	 5.002 & 	 0.813 & 	 0.184 \\
          ne &	71.02&	70.82&	71.71&	71.10& 	 8.525 & 	 0.693 & 	 0.187 \\
          chi&	68.31&	68.16&	68.07&	68.12& 	 2.502 & 	 0.037 & 	 0.070 \\
          kl&	69.62&	69.55&	70.49&	69.29& 	 11.227 & 	 0.418 & 	 0.361 \\
          wllr&	69.50&	69.96&	69.76&	69.19 & 	 2.155 & 	 0.218 & 	 0.048 \\
          orr&	69.49&	69.88&	69.50&	70.15& 	 4.970 & 	 0.560 & 	 0.168 \\
          dsidf&	71.74&	71.42&	72.02&	71.88& 	 7.870 & 	 0.490 & 	 0.139 \\
          dbidf&	70.61&	70.27&	70.43&	70.48& 	 6.401 & 	 0.641 & 	 0.122 \\
          rf&	70.49&	69.99&	70.43&	70.10& 	 6.400 & 	 0.177 & 	 0.166 \\
          cdm&	70.50&	70.69&	70.53&	70.19& 	 7.550 & 	 0.419 & 	 0.199 \\
          ngl&	69.50&	69.70&	69.53&	69.69& 	 4.821 & 	 0.519 & 	 0.080 \\
          cpd&	68.76&	69.22&	68.89&	68.84& 	 10.354 & 	 0.513 & 	 0.368 \\
    \hline
\end{tabular}

\caption{F1-score for each combination of global and local metrics on the Laptop dataset, the first four columns contain the F score of each combination but the last three ones contain the statistics of each global distribution.}
\end{center}
\end{table}
\begin{figure}[h!]
  \centering
   \includegraphics[width=\linewidth, height=12cm]{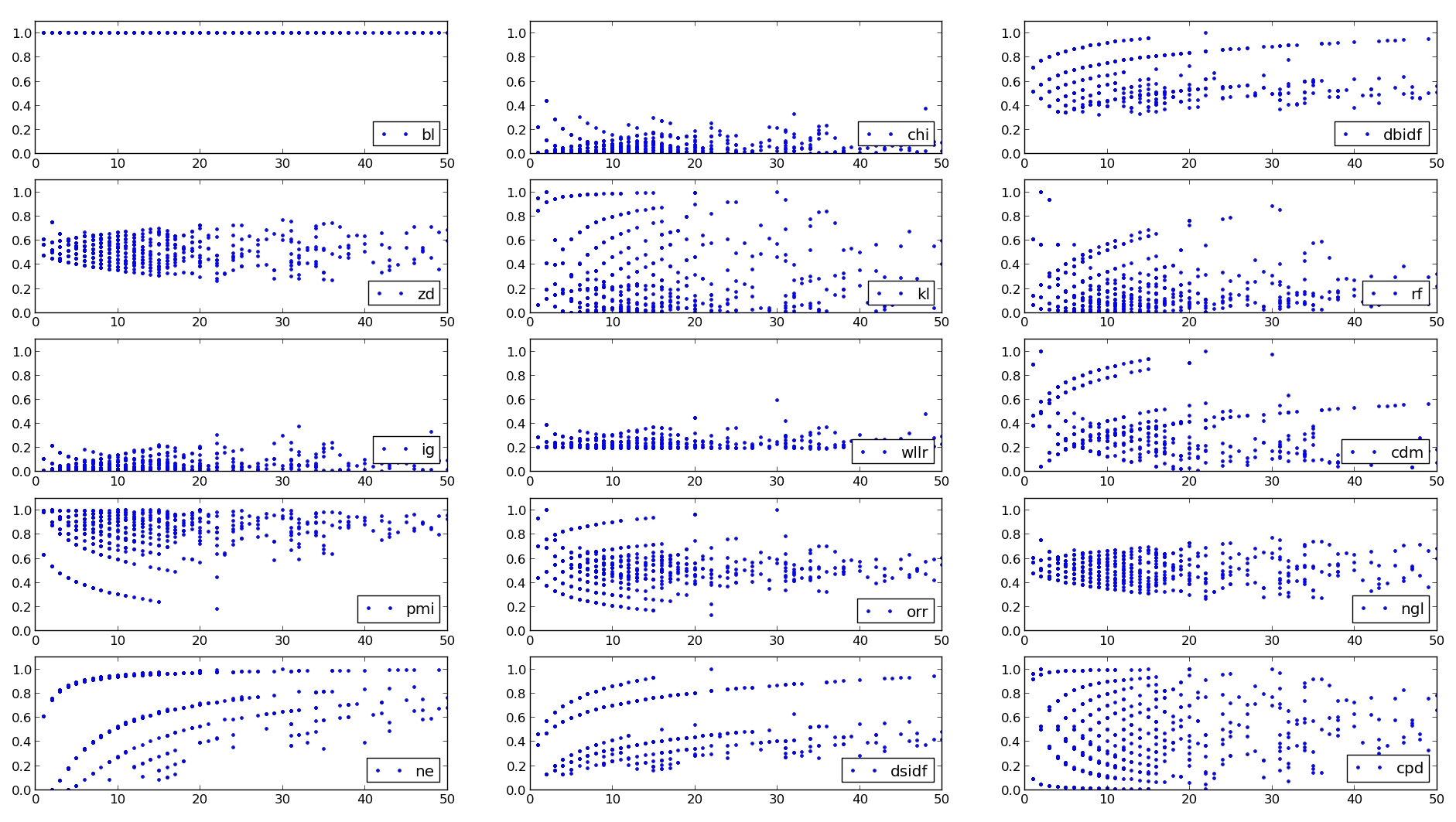}
  \caption{The distribution of each global metric in Laptop dataset.}
\end{figure}

Regarding local metrics, for the baseline, \textit{tf} produces the best result. For other metrics which improve the baseline, \textit{atf} often improves the \textit{tp} except with dbidf. \textit{logtf} always improves but the \textit{atf} weighting schema always gives the best results over all metrics except \textit{rf}. \textit{dsidf*atf and ne*atf} give the best results 72.02\%, 71.71\%, respectively.

Thus, most metrics improve the performance with \textit{tp} local weighting, and those who improve the \textit{bl*tp} produce the best results when using the \textit{tf} or \textit{atf} weighting.

Like for Twitter and restaurant reviews, Figure 3.3 shows the distribution of each metric over the laptop dataset.
\end{enumerate}
Thus, we can answer the questions of our research, can the global and local weighting schema improve the performance of a classifier? Obviously, the answer is yes, we point out that most global metrics improve the simple binary representation performance. We cannot conclude that there is always a metric which achieve the best result but it is more probably that some metrics always work well, the bad metrics are those who have narrow distribution and means close to zero. The local metrics seem to be influential in Twitter and Restaurant but not so influential in Laptop, we note that \textit{rf} is always among the best metrics over all the datasets.

\section{Conclusion and Future Work}
This study is an evaluation of supervised term weighting metrics for sentiment analysis in short text. We have studied fifteen different metrics of global weighting and four local weighting metrics. Three datasets are used for the evaluation. While our experimental results reveal that these metrics improve the polarity classification, there is no best choice for all datasets, several global metrics seem to work well with all datasets such as \textit{rf, kl, wllr}, and the local metrics often improve the performance especially \textit{tf} and \textit{atf}. We have analysed how each metric distributes the corpus in order to deduce the characteristics of the good and bad metrics. We have found that the bad metrics tend to have a narrow distribution with a mean value nearby to zero but we could not deduce some common characteristics among the metrics which give the best results.

In future work, we will investigate the normalization and the combination of term weighting metrics. Combining different metrics could be more effective than one, because it can use complementary information of the discriminating power of a term. And normalization can derease the difference in performance between the datasets with long and short document average length. 
\bibliography{bibigou1}

\begin{thebibliography}{}

\bibitem[Chen et~al., 2009]{chen_feature_2009}
Chen, J., Huang, H., Tian, S., and Qu, Y. (2009).
\newblock Feature selection for text classification with {Naïve} {Bayes}.
\newblock {\em Expert Systems with Applications}, 36(3, Part 1):5432--5435.

\bibitem[Church and Hanks, 1990]{church_word_1990}
Church, K.~W. and Hanks, P. (1990).
\newblock Word {Association} {Norms}, {Mutual} {Information}, and
  {Lexicography}.
\newblock {\em Comput. Linguist.}, 16(1):22--29.

\bibitem[Debole and Sebastiani, 2003]{debole_supervised_2003}
Debole, F. and Sebastiani, F. (2003).
\newblock Supervised {Term} {Weighting} for {Automated} {Text}
  {Categorization}.
\newblock In {\em Proceedings of the 2003 {ACM} {Symposium} on {Applied}
  {Computing}}, {SAC} '03, pages 784--788, New York, NY, USA. ACM.

\bibitem[Deng et~al., 2014]{deng_study_2014}
Deng, Z.-H., Luo, K.-H., and Yu, H.-L. (2014).
\newblock A study of supervised term weighting scheme for sentiment analysis.
\newblock {\em Expert Systems with Applications}, 41(7):3506 -- 3513.

\bibitem[Fan et~al., 2008]{fan_liblinear2008}
Fan, R.-E., Chang, K.-W., Hsieh, C.-J., Wang, X.-R., and Lin, C.-J. (2008).
\newblock {LIBLINEAR}: {A} {Library} for {Large} {Linear} {Classification}.
\newblock {\em Journal of Machine Learning Research}, 9:1871--1874.

\bibitem[Forman, 2003]{forman_extensive_2003}
Forman, G. (2003).
\newblock An {Extensive} {Empirical} {Study} of {Feature} {Selection} {Metrics}
  for {Text} {Classification}.
\newblock {\em Journal of Machine Learning Research}, 3:1289--1305.

\bibitem[Haddoud et~al., 2016]{haddoud_combining_2016}
Haddoud, M., Mokhtari, A., Lecroq, T., and Abdeddaïm, S. (2016).
\newblock Combining supervised term-weighting metrics for {SVM} text
  classification with extended term representation.
\newblock {\em Knowledge and Information Systems}, pages 1--23.

\bibitem[Hamdan et~al., 2014]{hamdan_impact_2014}
Hamdan, H., Bellot, P., and Bechet, F. (2014).
\newblock The {Impact} of {Z}\_score on {Twitter} {Sentiment} {Analysis}.
\newblock In {\em In {Proceedings} of the {Eighth} {International} {Workshop}
  on {Semantic} {Evaluation} ({SemEval} 2014)}, page 636.

\bibitem[Lan et~al., 2009]{lan_supervised_2009}
Lan, M., Tan, C.~L., Su, J., and Lu, Y. (2009).
\newblock Supervised and {Traditional} {Term} {Weighting} {Methods} for
  {Automatic} {Text} {Categorization}.
\newblock {\em IEEE Trans. Pattern Anal. Mach. Intell.}, 31(4):721--735.

\bibitem[Martineau and Finin, 2009]{martineau_delta_2009}
Martineau, J. and Finin, T. (2009).
\newblock Delta {TFIDF}: {An} {Improved} {Feature} {Space} for {Sentiment}
  {Analysis}.
\newblock In {\em {ICWSM}}, pages 10--15.

\bibitem[Nakov et~al., 2013]{nakov_semeval-2013_2013}
Nakov, P., Rosenthal, S., Kozareva, Z., Stoyanov, V., Ritter, A., and Wilson,
  T. (2013).
\newblock {SemEval}-2013 {Task} 2: {Sentiment} {Analysis} in {Twitter}.
\newblock In {\em Second {Joint} {Conference} on {Lexical} and {Computational}
  {Semantics} (*{SEM}), {Volume} 2: {Proceedings} of the {Seventh}
  {International} {Workshop} on {Semantic} {Evaluation} ({SemEval} 2013)},
  pages 312--320, Atlanta, Georgia, USA. Association for Computational
  Linguistics.

\bibitem[Ng et~al., 1997]{ng_feature_1997}
Ng, H.~T., Goh, W.~B., and Low, K.~L. (1997).
\newblock Feature {Selection}, {Perceptron} {Learning}, and a {Usability}
  {Case} {Study} for {Text} {Categorization}.
\newblock In {\em Proceedings of the 20th {Annual} {International} {ACM}
  {SIGIR} {Conference} on {Research} and {Development} in {Information}
  {Retrieval}}, {SIGIR} '97, pages 67--73, New York, NY, USA. ACM.

\bibitem[Nicholls and Song, 2010]{farzindar_comparison_2010}
Nicholls, C. and Song, F. (2010).
\newblock Comparison of {Feature} {Selection} {Methods} for {Sentiment}
  {Analysis}.
\newblock In Farzindar, A. and Kešelj, V., editors, {\em Advances in
  {Artificial} {Intelligence}}, volume 6085 of {\em Lecture {Notes} in
  {Computer} {Science}}, pages 286--289. Springer Berlin Heidelberg.

\bibitem[Nigam et~al., 2000]{nigam_text_2000}
Nigam, K., McCallum, A.~K., Thrun, S., and Mitchell, T. (2000).
\newblock Text {Classification} from {Labeled} and {Unlabeled} {Documents}
  {Using} {EM}.
\newblock {\em Machine Learning}, 39(2-3):103--134.

\bibitem[Paltoglou and Thelwall, 2010]{paltoglou_study_2010}
Paltoglou, G. and Thelwall, M. (2010).
\newblock A {Study} of {Information} {Retrieval} {Weighting} {Schemes} for
  {Sentiment} {Analysis}.
\newblock In {\em Proceedings of the 48th {Annual} {Meeting} of the
  {Association} for {Computational} {Linguistics}}, {ACL} '10, pages
  1386--1395, Stroudsburg, PA, USA. Association for Computational Linguistics.

\bibitem[Pang et~al., 2002]{pang_thumbs_2002}
Pang, B., Lee, L., and Vaithyanathan, S. (2002).
\newblock Thumbs {Up}?: {Sentiment} {Classification} {Using} {Machine}
  {Learning} {Techniques}.
\newblock In {\em Proceedings of the {ACL}-02 {Conference} on {Empirical}
  {Methods} in {Natural} {Language} {Processing} - {Volume} 10}, {EMNLP} '02,
  pages 79--86, Stroudsburg, PA, USA. Association for Computational
  Linguistics.

\bibitem[Pontiki et~al., 2015]{pontiki_semeval-2015_2015}
Pontiki, M., Galanis, D., Papageogiou, H., Manandhar, S., and Androutsopoulos,
  I. (2015).
\newblock {SemEval}-2015 {Task} 12: {Aspect} {Based} {Sentiment} {Analysis}.
\newblock In {\em In {Proceedings} of the 9th {International} {Workshop} on
  {Semantic} {Evaluation} ({SemEval} 2015)}, pages 486--495, Denver, Colorado.

\bibitem[Rehman et~al., 2015]{rehman_relative_2015}
Rehman, A., Javed, K., Babri, H.~A., and Saeed, M. (2015).
\newblock Relative discrimination criterion - {A} novel feature ranking method
  for text data.
\newblock {\em Expert Systems with Applications}, 42(7):3670 -- 3681.

\bibitem[Ren and Sohrab, 2013]{ren_class-indexing-based_2013}
Ren, F. and Sohrab, M.~G. (2013).
\newblock Class-indexing-based term weighting for automatic text
  classification.
\newblock {\em Information Sciences}, 236(0):109 -- 125.

\bibitem[Salton and Buckley, 1988]{salton_term-weighting_1988}
Salton, G. and Buckley, C. (1988).
\newblock Term-weighting approaches in automatic text retrieval.
\newblock {\em Information Processing \& Management}, 24(5):513 -- 523.

\bibitem[Savoy, 2012]{savoy_authorship_2012}
Savoy, J. (2012).
\newblock Authorship {Attribution} {Based} on {Specific} {Vocabulary}.
\newblock {\em Transactions on Information Systems (TOIS)}, 30(2):1--30.

\bibitem[Savoy, 2013]{savoy_feature_2013}
Savoy, J. (2013).
\newblock Feature {Selections} for {Authorship} {Attribution}.
\newblock In {\em Proceedings of the 28th {Annual} {ACM} {Symposium} on
  {Applied} {Computing}}, {SAC} '13, pages 939--941, New York, NY, USA. ACM.

\bibitem[Sebastiani, 2002]{sebastiani_machine_2002}
Sebastiani, F. (2002).
\newblock Machine {Learning} in {Automated} {Text} {Categorization}.
\newblock {\em Computing Surveys (CSUR)}, 34(1):1--47.

\bibitem[Shaw, 1995]{shaw_term-relevance_1995}
Shaw, Jr., W.~M. (1995).
\newblock Term-relevance {Computations} and {Perfect} {Retrieval}
  {Performance}.
\newblock {\em IInformation Processing and Management}, 31(4):491--498.

\bibitem[Simeon and Hilderman, 2008]{simeon_categorical_2008}
Simeon, M. and Hilderman, R. (2008).
\newblock Categorical {Proportional} {Difference}: {A} {Feature} {Selection}
  {Method} for {Text} {Categorization}.
\newblock In {\em Proceedings of the 7th {Australasian} {Data} {Mining}
  {Conference} - {Volume} 87}, {AusDM} '08, pages 201--208, Darlinghurst,
  Australia, Australia. Australian Computer Society, Inc.

\bibitem[Wu and Gu, 2014]{wu_reducing_2014}
Wu, H. and Gu, X. (2014).
\newblock Reducing {Over}-{Weighting} in {Supervised} {Term} {Weighting} for
  {Sentiment} {Analysis}.
\newblock In {\em {COLING} 2014, 25th {International} {Conference} on
  {Computational} {Linguistics}, {Proceedings} of the {Conference}: {Technical}
  {Papers}, {August} 23-29, 2014, {Dublin}, {Ireland}}, pages 1322--1330.

\bibitem[Yang and Pedersen, 1997]{yang_comparative_1997}
Yang, Y. and Pedersen, J.~O. (1997).
\newblock A {Comparative} {Study} on {Feature} {Selection} in {Text}
  {Categorization}.
\newblock In {\em Proceedings of the {Fourteenth} {International} {Conference}
  on {Machine} {Learning}}, {ICML} '97, pages 412--420, San Francisco, CA, USA.
  Morgan Kaufmann Publishers Inc.

\end{thebibliography}
\bibliographystyle{apalike}
\end{document}